# A fully data-driven method to identify (correlated) changes in diachronic corpora

Alexander Koplenig

Institute for the German language (IDS), Mannheim, Germany

**Abstract**

In this paper, a method for measuring synchronic corpus (dis-)similarity put forward by Kilgarriff (2001) is adapted and extended to identify trends and correlated changes in diachronic text data, using the Corpus of Historical American English (Davies 2010a) and the Google Ngram Corpora (Michel et al. 2010a). This paper shows that this fully data-driven method, which extracts word types that have undergone the most pronounced change in frequency in a given period of time, is computationally very cheap and that it allows interpretations of diachronic trends that are both intuitively plausible and motivated from the perspective of information theory. Furthermore, it demonstrates that the method is able to identify correlated linguistic changes and diachronic shifts that can be linked to historical events. Finally, it can help to improve diachronic POS tagging and complement existing NLP approaches. This indicates that the approach can facilitate an improved understanding of diachronic processes in language change.

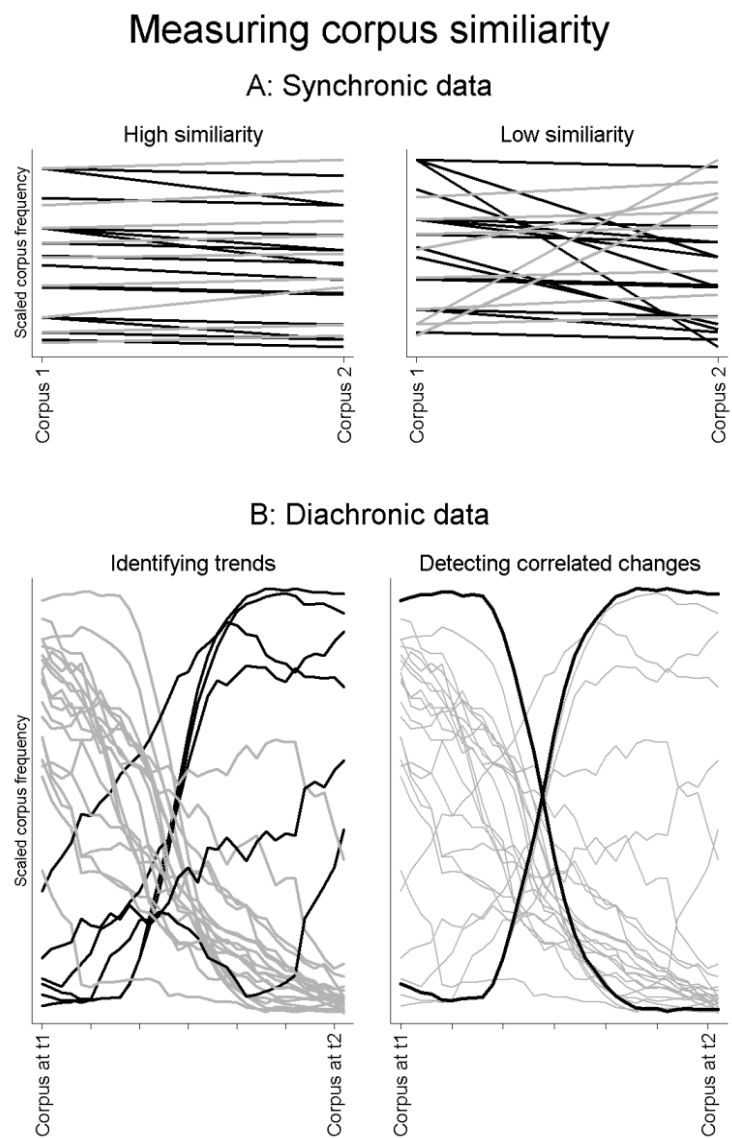

**Figure 1: Abstract visualization of the method adapted in this paper.**

## 1. Introduction

'The scientific value of heuristic statistical methods may be illustrated by a metaphor: You will see different things if you walk, ride a bike, go by car, or look down from a plane. Statistics is a vehicle which can be used at arbitrary "velocity" and arbitrary "height", depending on how much overview you wish for and how detailed you want to look at a linguistic "landscape".' (Köhler 2012: 16)

The recent availability of large machine-readable diachronic corpora, such as the Helsinki Corpus (Rissanen et al. 1991; Kohnen 2007), the Old Bailey corpus (Huber et al. 2012), the TIME magazine corpus (Davies 2007), the COHA (Davies 2010a; Davies 2012) or the Google N-gram corpora (Michel et al. 2010a; Davies 2014), affords novel possibilities to analyze and understand linguistic change. A linguist interested in the diachronic development of one linguistic structure relative to another one, for example, could use one of the aforementioned corpora, extract the time series of the respective forms and interpret potential frequency changes across time using the statistical method of time series analysis in general (Becketti 2013), or the methods specifically developed for historical corpus linguistics (Hilpert and Gries 2009). To a certain extent, however, this approach presupposes that the researcher already knows which particular linguistic forms to begin with, as the examples of Hilpert and Gries (2009: 386–388) imply. One drawback to this approach is the fact that potentially interesting trends and correlated processes in diachronic text data could remain undetected given the sheer amount of available data (Bentz et al. 2014: 2).

Against this backdrop, this paper presents a fully data-driven method, which automatically extracts word types that have undergone the most pronounced change in frequency in a given period of time. This method, first developed by Kilgarriff (2001; 2012) to compare synchronic corpora, combined with statistical methods from time series analysis finds meaningful patterns and relationships in diachronic corpora. In addition, it can help to improve diachronic POS tagging and complement existing NLP approaches. This indicates that the approach as can facilitate an improved understanding of diachronic processes in language change and, to rephrase Köhler's metaphor quoted above, can be used in this way as 'a vehicle to give an overview of a (diachronic) linguistic landscape'.

The remainder of this paper is structured as follows: First, the data sets used in this study are presented (Section 2). In the next section, a strategy to assess synchronic corpus similarity is introduced (Section 3). This section is followed by a case study that shows how the method can help to visualize differences between synchronic corpora (Section 4). In Section 0, the method is adapted to the analysis of diachronic corpora, followed by two case studies (Section 6 and Section 7) that demonstrate that the method makes it possible to identify diachronic trends and detect correlated changes. This paper ends with some concluding remarks discussing how the method could be further improved and extended (Section 8). Further material is available online and as supplemental online material.

## 2. Data

Corpus data from three different sources are used in this paper.

(A) The first resource is a set of three unlemmatised frequency lists compiled by Kilgarriff (1997) on the basis of the British National Corpus (BNC). One represents the written part, which comprises 89.7M tokens. The other two frequency lists represent the spoken part of the BNC. The first of these covers 'context-governed' spoken material (e.g. lectures, meetings, news commentaries) and consists of 6.2M tokens. The second list covers 'demographic' spoken material (i.e. recorded daily conversations) and consists of 4.2M tokens. The lists are freely available online (Kilgarriff 2014).

(B) Secondly, I use unigram data from the Corpus of Historical American English (COHA), which contains 400M tokens from the period between 1810 and 2009. It contains all unique word strings that occur at least three times in total. The data are freely available online (Davies 2010a).

(C) Thirdly, I use unigram and bigram data from the Google books Ngram corpora (GBC), made available by Michel et al. (2010a). For this study, the datasets of Version 2 (July 2012) of the following languages were used (including two varieties of English): American English, British English, French, German, Italian and Spanish. All unigram corpora share the same basic structure, in which the first column is the string variable for the word, the second variable contains the word-class (POS) information as described in Lin et al. (2012) and the third column contains the match count for one particular year (e.g. match1899). For the bigram corpora, the structure is similar, comprising a string variable and word class information both for the first and for the second word in one bigram. The data contains all unique word strings that occur at least 40 times in total. The data are freely available online (Culturomics 2014).

Corpus size varies considerably for the synchronic data (A) and strongly increases as a function of time for the diachronic data (B and C). To avoid a potential systematic bias (Tweedie and Baayen 1998), an efficient and computationally cheap solution is to draw random samples of 1,000,000 tokens from the data by performing a binomial split for each corpus (as suggested by Piantadosi 2014). For each word type *w*, this procedure returns binomial ($n_{wc}$, $p_c$) random variates, where $n_{wc}$ is the raw token frequency of the word type *w* in the corpus *c* and *p* is the probability of success, which is given as: $(1{,}000{,}000 + 10{,}000)/N_c$, where $N_c$ is the corpus size of corpus *c*.[1] The resulting corpora of 1,000,000 tokens are what Tweedie & Baayen (1998) would call fully randomised samples of all texts in a given corpus or in a given year.[2]

All analyses in this paper were carried out using Stata/MP2 12.1 for Windows (64-bit version). All Stata do-files are available upon request.

### 3. Measuring similarity in synchrony

In this section, the mathematical background of the method to measure (dis-)similarity in synchrony, which will be adapted and extended to measure similarity in diachrony in the following sections, is discussed and put into perspective in regard to other measures of statistical divergence.

Kilgarriff (2001) discusses and evaluates several methods for the measurement of synchronic corpus similarity. All measures only need frequency vectors as input, using the token frequency of each orthographic word form. To compare two different corpora *c*1 and *c*2, the individual token frequencies are first aggregated to generate a new corpus *u*. For his analysis, Kilgarriff (2001: 253) then uses the 500 most frequent words of *u*, but notes that this is done for convenience only, there is no statistical reason for this choice. For the analyses in this paper, all words in *u* are used to calculate the similarity measure. In the experiment presented below, I will demonstrate that the correlation between a similarity measure based on the top 500 words and a measure based on all words is generally very high.

Kilgarriff (2001) works out a $\chi^2$-based measure. In his analysis, this measure works best for the comparison of different corpora. For each word $w_i$ and each corpus $c$, an expected frequency $e_i$ based on the union of the two corpora $u$ is calculated using the following formula:

$$e_{i,c} = p_{i,u} \cdot N_c \qquad (1)$$

where $p_{i,u}$ is the relative frequency of word $i$ in the union of the two corpora $u$ and $N_c$ is the corpus size of corpus $c$. Then the partial contribution $\chi^2_{i,c}$ to the 'grand total' similarity $\chi^2$ of word $w_{i,c}$ with an observed frequency $o_{i,c}$ in corpus $c$ is defined as:

$$\chi^2_{i,c} = \frac{(o_{i,c} - e_{i,c})^2}{e_{i,c}} \qquad (2)$$

In this paper, the partial value will be used to identify the words that are most important for measuring synchronic and especially diachronic corpus similarity.

The similarity $\chi^2$ between two corpora is then just the sum of all partial values where $v$ is the vocabulary size of $u$.

$$\chi^2 = \sum_{i=1}^{v} \chi^2_{i,c} \qquad (3)$$

The rationale of this procedure is very intuitive: if $c_1$ and $c_2$ are very similar, then the distribution of token frequencies should also– apart from random fluctuations – be very similar. Therefore, the individual deviations from the expected frequencies as calculated in (1) will tend to be very small. If, for example, two identical frequency lists are being compared, $\chi^2$ will be zero. If however $c_1$ and $c_2$ are very dissimilar, the expected frequencies for some words will be quite different from the observed frequencies and the squared sum of those differences will be very high. The two plots of (A) in Figure 1 visualize this idea: To compare two different corpora, the frequency differences between the word types are used as a proxy for general similarity. As can be seen on the left side of plot A, the distributions of token frequencies are – apart from random fluctuations – very similar, the lines are almost parallel. The right side of plot A shows that if $c_1$ and $c_2$ are very dissimilar, the expected frequencies of some words are quite different from their observed frequencies, and the squared sums of those differences are very high. For diachronic data (plot B), one corpus is analyzed at different moments in time. Thus, the method helps to identify the word types that underwent the most pronounced changes in frequency (left side). In addition to that, the method can be used to detect correlated changes (right side). This idea is developed in more detail in Section 0.

Illustrative examples of the method can also be found in Kilgarriff and Salkie (1996). Kilgarriff (2001: 255) notes that it is a desirable consequence of the approach that differences for higher-frequency words are more important in determining their individual contribution to the (dis-)similarity. If, for instance, the determiner 'the' has a very different frequency in two corpus samples, then this signals a stronger dissimilarity compared to a frequency difference for a relatively infrequent element.

Given the fact that the measure is not scale-independent (Kilgarriff 2001: 258), Cramér's *V* (Cramér 1946: 282) is used instead of Kilgarriff's original approach. For the comparision of two corpora, *V* is given as:

$$V = \sqrt{\frac{\chi^2}{N_u}} \qquad (4)$$

where $N_u$ is the corpus size of the union of the two corpora $u$. Cramér's *V* is a classical measure of the strength of association, ranging from 0 for no association to 1 for a very strong association. For the comparison of two corpora, small values signal a very high similiarity, while higher values are indicative of disimilar corpora.

Compared to other measures used in his analysis, Kilgarriff (2001: 258) notes that the $\chi^2$-based measure is 'not rooted in a mathematical formalism' and identifies this as an area for future research. However, as a study by Endres and Schindelin (2003) suggests, the measure is, if all words are used to calculate the similarity, approximately equivalent to the Jensen-Shannon divergence (JSD), which can be defined as (Klingenstein et al. 2014):

$$JSD(\vec{c_1}||\vec{c_2}) = 0.5 \left[ KL\left(\vec{c_1} \middle| \frac{\vec{c_1}+\vec{c_2}}{2}\right) + KL\left(\vec{c_2} \middle| \frac{\vec{c_1}+\vec{c_2}}{2}\right) \right] \qquad (5)$$

where $\vec{c_1}$ and $\vec{c_2}$ are the two corpus frequency vectors and *KL* is the Kullback-Leibler divergence, which is given as:

$$KL(P||Q) = \sum_x P(x) \log \frac{P(x)}{Q(x)} \qquad (6)$$

where *P(x)* and *Q(x)* are the probability distributions of the two vectors. When *P(x) = 0* but *Q(x) ≠ 0*, the KL-divergence is undefined, since the logarithm of zero is also undefined. For the analysis presented below (cf. Figure 2), it is stipulated that those cases are interpreted as zero in the calculation of the sum in (6), because $\lim_{x \to 0} x \log(x) = 0$. Bochkarev et al. (2014) demonstrate that more principled but also more complicated approaches in this context do not lead to qualitatively different results. In practice, the Jensen-Shannon divergence measures the divergence of each distribution from the mean of the two distributions (Jurafsky and Martin 2009: 700). It has already been fruitfully employed in the context of measuring stylistic influences in the evolution of literature (Hughes et al. 2012), cultural and institutional changes (Klingenstein et al. 2014), or the dynamics of lexical evolution (Bochkarev et al. 2014). Pechenick et al. (2015) use the JSD measure to quantify the changing compositions of the English GBC set with a particular attention to word types with a high partial contribution to the observed divergence (cf. Section 3 and Section 0).

To show that both $V$ and the square root of the *JSD* are strongly correlated, I sampled two million tokens from the written part of the BNC (cf. Section 2). The resulting frequency list is used as the union corpus $u$. For each word type $w$, a binomial split was performed, where $n_{w,c}$ is the raw token frequency of the word type $w$ in $u$ and $p$ is the probability of success. The resulting number of tokens $f$ then represents the token frequency of $w$ in $c_1$, while the difference between the total token frequency in $u$ minus $f$ represents the token frequency of $w$ in $c_2$. The success probability $p$ was randomly varied over the interval $0.5\pm r$. To generate pairs of corpora with varying degrees of similarity, $r$ was gradually incremented from 0.004 to 0.4. Corpus pairs with a small $r$ should have a greater similarity than corpus pairs with a bigger $r$, because – on average – half of the tokens of word type $w$ in $u$ are classified as belonging to $c_1$ and half of the tokens as belonging to $c_2$. With an increasing $r$ however, the probability of success for each word type varies more strongly around 0.5. So, for some word types, more tokens are placed into $c_1$, while for other word types, most tokens are put into $c_2$. On average, the corpus size for each corpus remained approximately equal ($\approx$ 1M tokens).

Using this technique, 1,000 pairs of corpora were generated and the *JSD* and the $\chi^2$-value for the first 500 words as suggested by Kilgarriff (2001: 253) were calculated for all pairs, $V$. Figure 2 demonstrates that all three measures seem to work as intended: corpus pairs with smaller values of $r$ (indicated by darker shades of gray) are classified as being more similar (indicated by a small *JSD*, a small *V* and a small $\chi^2$-value). Corpus pairs with higher values of *r* (indicated by lighter shades of gray), on the other hand, are classified as being less similar.

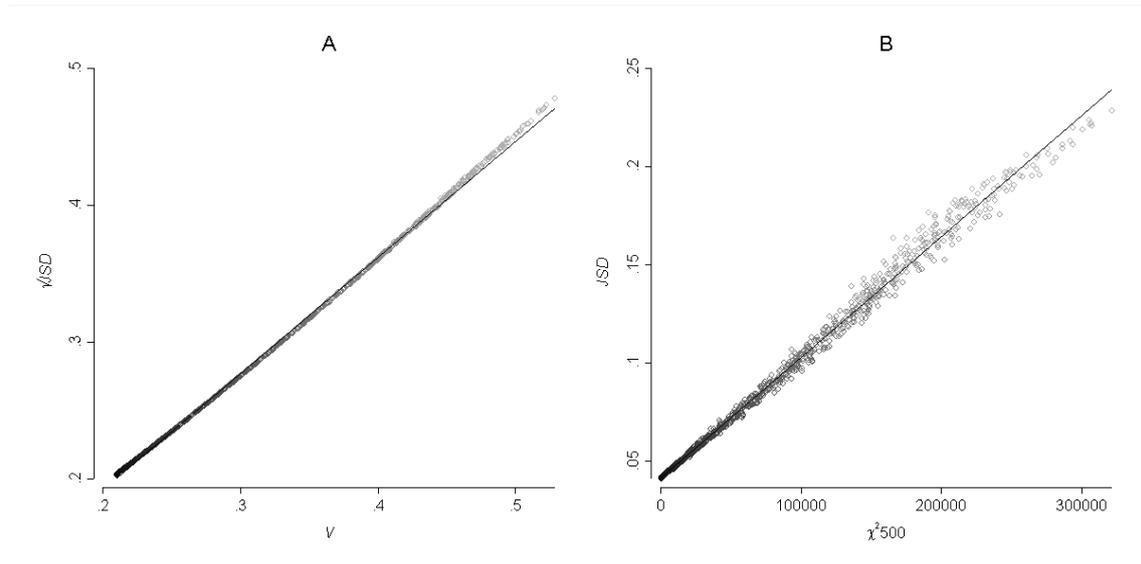

**Figure 2: Measuring the distance between corpora with varying degrees of similarity (darker shades of gray indicate more similar corpora). Solid black lines: linear fit between the observed values (hollow circles).**

In addition to that, Plot A of Figure 2 shows that the square root of *JSD* and *V* are strongly correlated ($\rho = 0.9996$) and approximately equivalent. Plot B of Figure 2 demonstrates that the $\chi^2$-value for the first 500 words and the *JSD* are also strongly related ($\rho = 0.9972$). For all subsequent analyses, all words are used to calculate the similarity between two corpora as mentioned above.

## 4. Case study 1: Visualizing the difference between the two spoken parts of the BNC

Although the main focus of this paper is on the analysis of diachronic data, a synchronic application can be useful to illustrate the potential of the method. For this purpose, I first measured the similarity between one million token samples of the written and the two spoken parts of the BNC (cf. Section 2). The greatest similarity is calculated for the two spoken parts ($V = 0.384$). It is also in line with *a priori* expectations that the demographically sampled spoken part (*demog*) is less similar to the written sample ($V = 0.598$), compared to the context-governed part (*cg*) of the BNC ($V = 0.448$), because the *demog* part primarily consists of informal English conversations, while the *cg* part incorporates more informative spoken material (Burnard 2007).

In this context, one might be interested in the main differences between these two parts of the BNC. Using formula (2) to measure the partial contribution for each word type, Figure 3 plots the elements with the most pronounced differences, showing 64 function words and 64 content words.[3] The size of the words is proportional to the (log of the) partial contribution of the respective word type. Lighter shades of gray indicate that the word type is more frequent in the *demog* sample, while word types in black are more frequent in the *cg* sample.

For example the determiner 'the' is the word type with the highest partial contribution ($\chi^2 = 3276.75$). It has a *cg*-token frequency of 47,592 and a much lower *demog* frequency of 27,421. In reference to a classical study on genre differences (Biber and Finegan 1989), the visualization fits nicely with the fact that the *demog* part is dominated by daily conversations which can be characterized by linguistic features that are typical for interactive or involved text production, e.g. present tense verbs, contractions ('ll', 'm', 's', 'n't'), 1st- & 2nd-person pronouns (except 'we') or the pronoun 'it'. On the other side of this continuum, the *cg* part of the BNC contains more prepositions (e.g. 'in', 'to', 'into', 'at', 'by', 'from') and more nouns (as indicated by the higher frequency of the determiners 'the' and 'a'), which are linguistic features that signal informational text production (Biber and Finegan 1989: 490–492). Accordingly, an additional analysis reveals that the *cg* sample contains 15,970 different nouns, while the *demog* sample only contains 13,052 nouns. The *written* sample even contains 36,072 nouns (regular expression to count the number of nouns: nn*). Further analyses focusing on different word classes are possible using this approach.

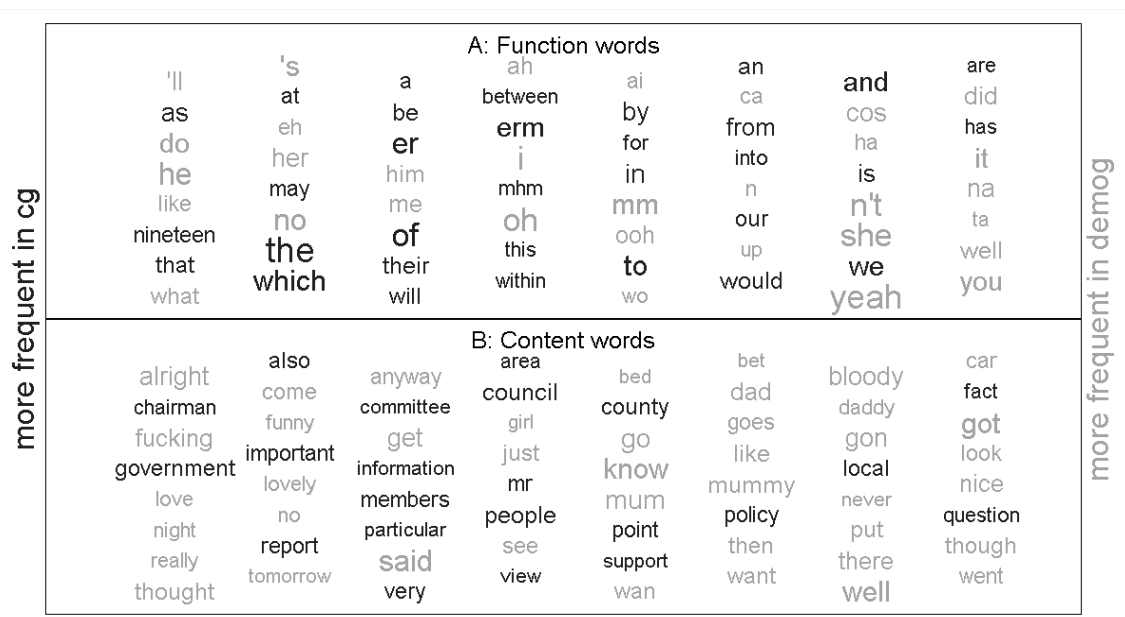

**Figure 3: Visualization of the differences between the two spoken-parts of the BNC. The size of the words is proportional to the (log of the) partial contribution of the respective word type. Lighter shades of gray indicate that the word type is more frequent in the demog sample, while black word types are more frequent in the cg sample.**

## 5. Measuring corpus similarity in diachrony

The illustration presented in the last section showed how the Kilgarriff approach (2001) can be used to fruitfully measure and visualize differences between synchronic data. Since a diachronic corpus is basically just 'a collection of texts that vary along the parameter of time' (Gries and Hilpert 2008: 386), the focus will now move on to the analysis of diachronic data, using the same methodology (cf. Mota 2010 for a similar approach in this direction). Instead of comparing two different corpora, synchronic snapshots (i.e. word type vectors of token frequencies) of one diachronic corpus are compared across different successive moments in time. In this case, the similarity *V* between time point $t_1$ and $t_2$ is the square root of the sum of all partial values divided by corpus size of the union of the two synchronic snapshots. Using this methodology, small values of *V* indicate that in the investigated period of time, no pronounced lexical changes have taken place. If however, the frequencies of many word types have changed, the expected frequencies of those words will strongly diverge from the observed frequencies, and as a consequence, the value of *V* will be bigger.

Compared to the synchronic case (cf. plot A of Figure 1), using formula (2) to measure the partial contribution for each word type has a natural interpretation for diachronic data (cf. the left side of plot B of Figure 1): large partial values indicate that the respective word types underwent pronounced changes in frequency. In terms of time series analysis, this means that word types with a large partial value have a strong upward or downward trend. This property allows us to adopt a new analytical perspective, because the method can be used to detect correlated changes (cf. the right side of plot B of Figure 1), that is, it can find word types whose frequency changes are positively or negatively correlated (e.g. word types that have potentially replaced each other). To calculate correlations between time series of different word types, a few important notes are in order about the special properties of time series analysis. In another publication, I have discussed this in more detail (Koplenig 2015b). The statistical analysis of time series is special; mainly because there is basically no such thing as a univariate time series: 'So-called univariate time-series analysis actually is the analysis of the bivariate relationship between the variable of interest and time.' (Becketti 2013: 92). If we use the method outline above, then by definition, the word types with the strongest trend will be extracted. In time series analysis, those time series are said to be non-stationary. If two such series are correlated with each other, those time series will, by mathematical necessity, always look highly correlated when in fact they are not related in any substantial sense (Granger and Newbold 1974). Why is that? Basically, the Pearson product-moment correlation coefficient is the covariance of two variables *x* and *y* scaled to the interval [-1,1] (a similar logic applies for other related normalized variants of the inner product such as the cosine, cf. Jurafsky and Martin 2009: 697–701). The covariance measures whether values of *x* that are above/below average tend to co-occur with values of *y* that are above/below average. Now, if we pre-select trending time series for our comparisons, the diachronic values of those time series will spuriously correlate, either positively or negatively. Thus, the correlation coefficient for those two time series will be artificially high, even if the two series are completely unrelated.

This problem can be solved by correlating period-to-period changes instead of comparing actual values of the time series (for a visualization, see Koplenig 2015b; or Koplenig 2015c). The rationale of this procedure is simple: if we compare the differences of two time series *x* and *y*, a strong positive correlation implies that period-to-period changes that are above/below the average for *x* correspond mainly to changes that are above/below the average for *y*. For each time series *x*, the difference between a year and five years before is calculated using the following notation:

$$\Delta x_t = x_t - x_{t-5} \qquad (7)$$

To detect correlated changes, the resulting differences are then compared for the time series of two word types in a given period of time. The choice of the five year time span is somewhat arbitrary. In principle, other period-to-period spans are possible.

In what follows, two case studies will demonstrate that the method is useful for the identification of lexical changes and short term diachronic shifts that can be linked to historical events. In addition to that, it might help to improve diachronic POS tagging, and it might complement other NLP approaches. This indicates that the approach can be fruitfully applied to the analysis of diachronic processes in language change.

## 6. Case study 2: Lexical changes in the COHA

Compared to the GBC, the COHA is balanced in regard to both genre and sub-genre across decades. It is hoped that this 'allows researchers to examine changes and be reasonably certain that the data reflects actual changes in the "real world", rather than just being artifacts of a changing genre balance' (Davies 2010a). Using the information about the composition of the COHA (Davies 2010b), plot A of Figure 4 shows that the ratio of fiction is closely around 50% for each decade (48 – 55%). Plot B of Figure 4 demonstrates that the ratio of dramatic fiction varies between zero and roughly ten percent (in 1810). To measure similarity between decades, *V* is calculated as described above. The solid line in plot C of Figure 4 plots the value of *V* between the decade *d* and *d*+10 for all decades (left y-axis), while the dashed line plots *V* excluding the 1810s (right y-axis), since the large difference between the 1810s and the 1820s would mask subsequent processes. Comparing plot B and plot C demonstrates that *V* seems to capture the higher fraction of dramatic fiction in the 1810s (solid line). In addition, the dashed line also shows that *V* is sensitive to the fact that the ratio of dramatic fiction rises around the 1900s and remains at a similar level afterwards. The contour plot D of Figure 4 shows *V* between all pairs of decades.

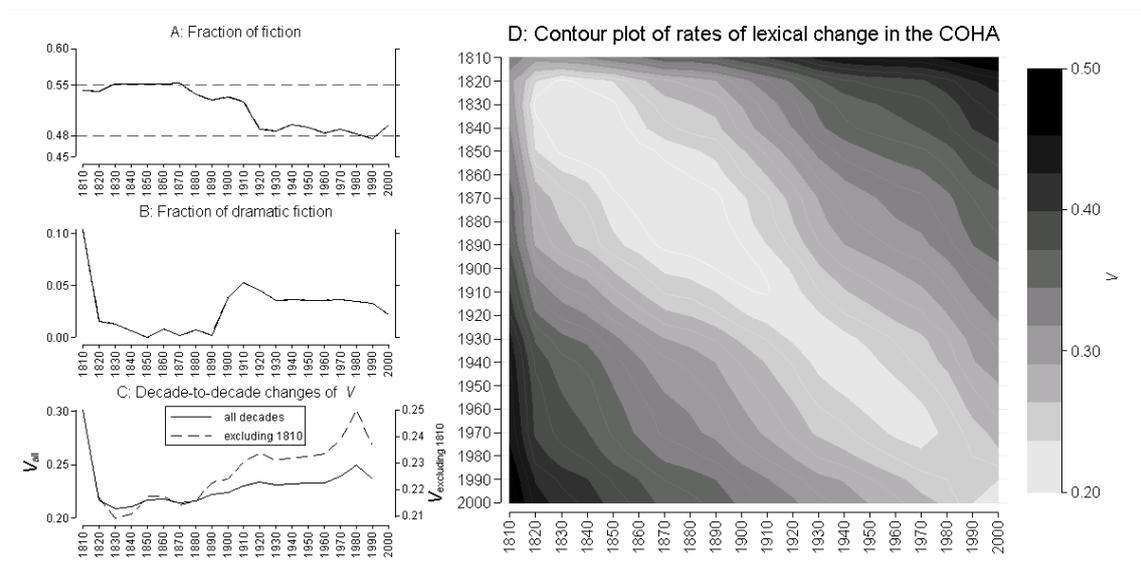

**Figure 4: Visualizing changing genre compositions of the COHA using the *V*-method.**

One might object that it is not clear whether the observed changes can really be attributed to the changing ratio of dramatic fiction (cf. Plot C of Figure 4). To refute this objection, Figure 5 plots the 64 most different word types between the 1810s and the 1820s. Additionally, time series for the 20 word types with the highest partial contribution are visualized as small multiples (Tufte 2001). Figure 5 clearly shows that the higher ratio of dramatic fiction for the 1810s explains the observed dissimilarity between this decade and the other decades. Words like 'Enter', 'Exit', or 'SCENE' and proper nouns in general are typical for dramas.

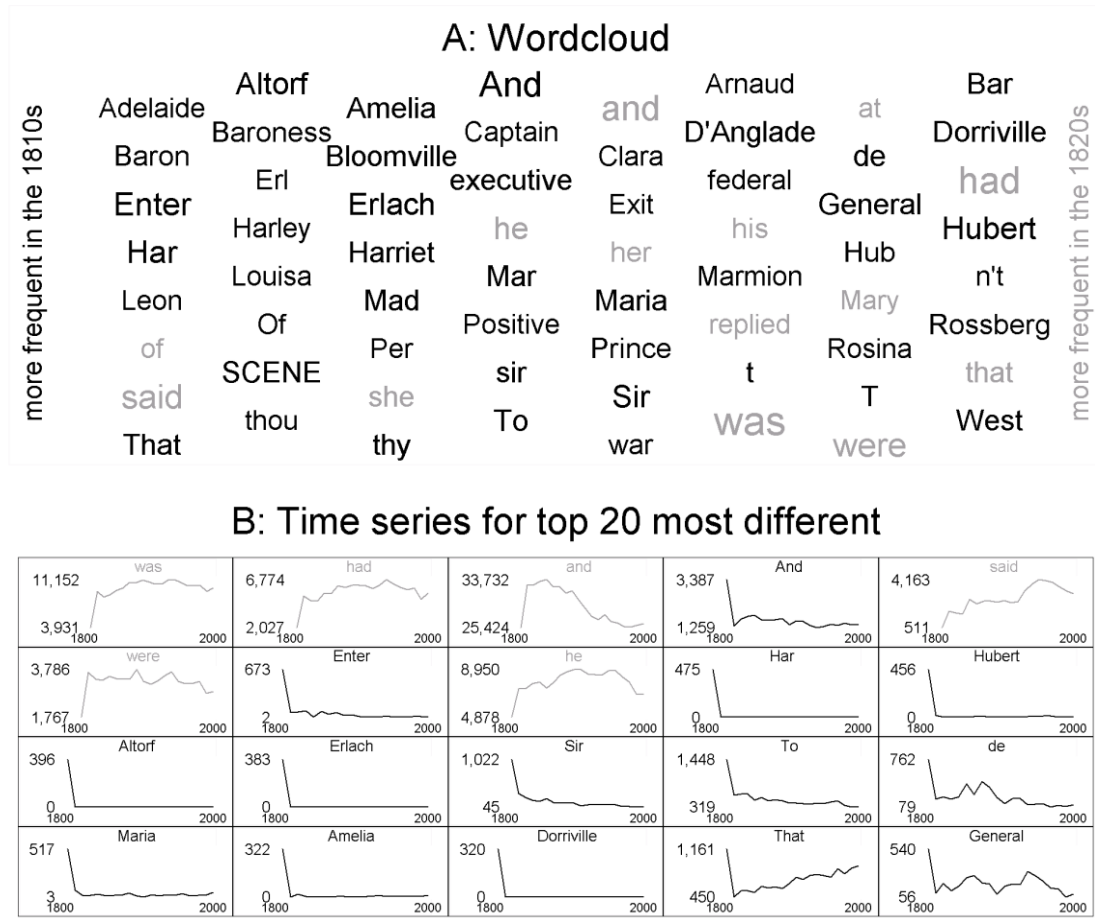

**Figure 5: Visualization of the differences between the 1810s and the 1820 in the COHA. The size of the words is proportional to the (log of the) partial contribution of the respective word type. Lighter shades of gray indicate that the word type is more frequent in the 1820s, while black colored word types are more frequent in the 1810s. The labels on the y-axes of the small multiples show maximum and minimum values in the 1M token samples (cf. Section 2).**

Using the information about the composition of the COHA further reveals that 'Marmion' is a play by J. N. Barker; 'Hub' and 'Mar' are abbreviations of names (Hubert & Marcello) in a play by I. Harby called the Gordian Knot; Harriet 'Har' Bloomville is one of the main characters in 'The Poor Lodger' by W. C. White and if one reads 'The Tooth-Ache' by B. John, one finds out the 'de' is used to dramatize a strong French accent as in 'dis is de same vay, avec tout le monde […] you see de pain go avay for little […]' (p. 26).

To sum up this section, the analysis indicates that the method outlined above can help to automatically identify and understand changing genre compositions in diachronic corpora.

### 7. Case study 3: Rates of lexical change in the Google Books Ngram corpora

In general, $V$ can be viewed as approximating the 'rate of lexical change' (Bochkarev et al. 2014: 5) in a given diachronic corpus and a given timespan. To illustrate this, Figure 6 plots those rates as represented in the GBC data for six languages including two varieties of English (American English, British English, French, German, Italian and Spanish) for each year in the timespan between 1800-2009

For each language, the rate of change is calculated by comparing the word frequency vector of a year with the word frequency vector of five years before that year (e.g. 1820 is compared with 1815). Since, especially for Italian and Spanish, $V$ was much higher in the $19^{th}$ compared to the $20^{th}$ century, probably due to the small size of the GBC in the $19^{th}$ century, potentially interesting processes of change in the $20^{th}$ century would be masked. Therefore, a scale break was added between the $19^{th}$ and the $20^{th}$ century (Cox 2012). This means that the time series for each language are divided into two panels, one for the 19th century (left y-axis) and one for the 20th century (right y-axis).

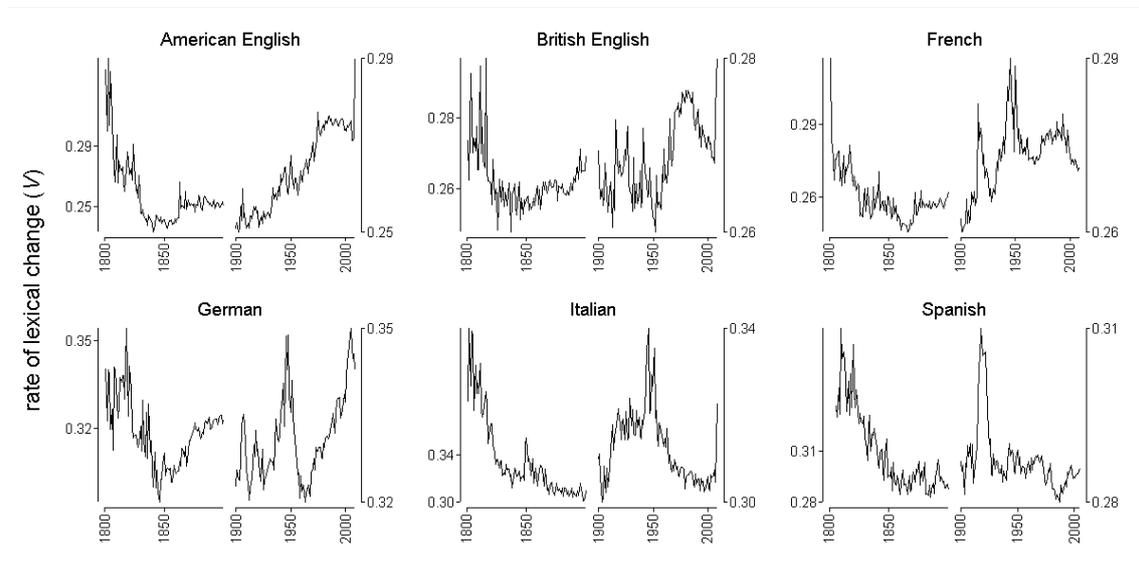

**Figure 6: Lexical changes in the Google Books Ngram corpora.** *Note*: a scale break was added around 1900; thus the y-axis on the left shows the time series from 1800 to 1899, while the y-axis on the right shows the series from 1900 to 2009.

A striking aspect in Figure 6 is the fact that there are visible spikes in the period of the two world wars for all investigated languages. In the following, I want to demonstrate how the method of measuring corpus similarity can help to understand what drives those changes. Firstly, it will be shown that the spike around 1910 for the Spanish data has nothing to do with the World War I, but with orthographic changes. In this context, the method might be valuable for the improvement of diachronic POS tagging and to find variants of word forms that might have replaced each other. Secondly, regarding the spikes around the same time for the German data, it will be shown that there are mainly two reasons: an orthographic reform and lexical changes that can be linked to contingent historical events, in this case WWI.

**Spanish Google Books Ngram lexical changes around 1910**

To understand the sudden change around 1910 for Spanish data (cf. Figure 6), I used a similar approach as before in Figure 5. Compared to the COHA corpus, the GBC is much bigger and has its data time-stamped year by year instead of decade by decade.

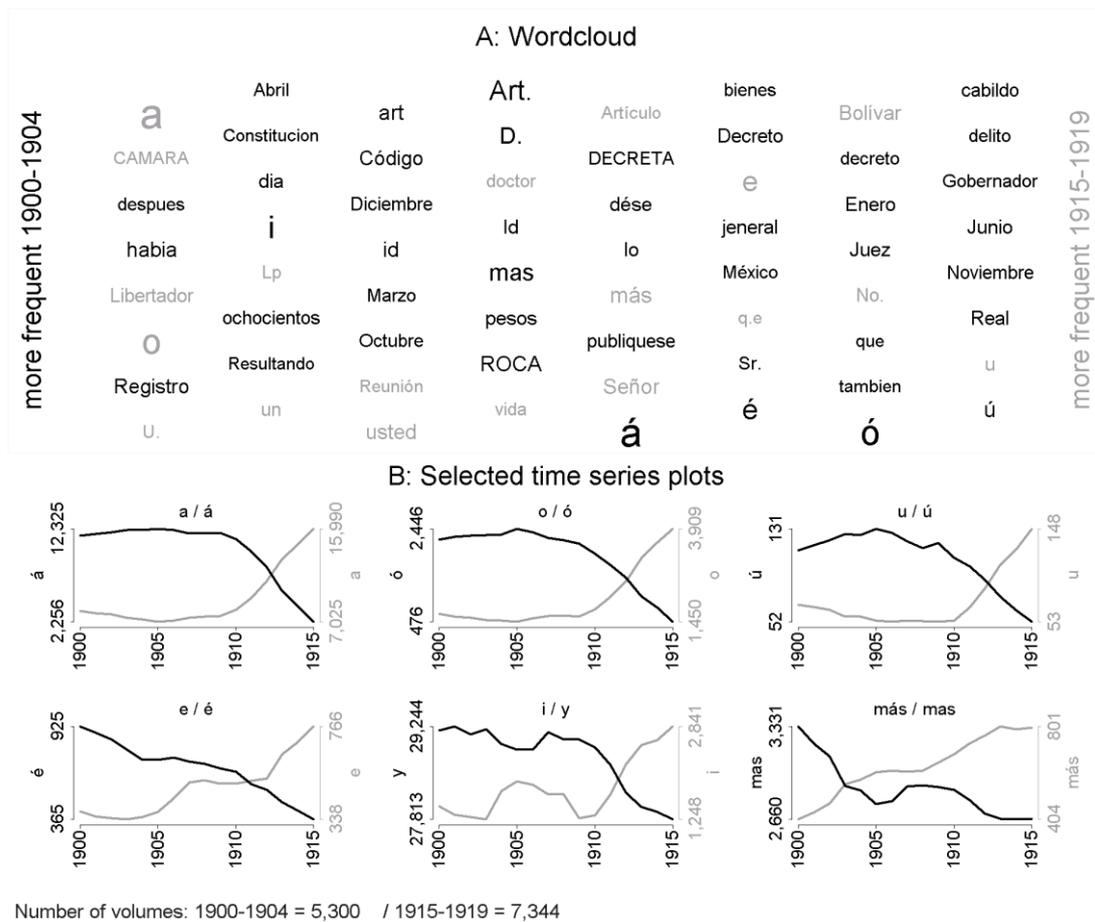

Figure 7: Visualization of the differences between the time spans 1900-1904 and 1915-1919 in the Spanish Google Books Ngram data (cf. Figure 5 for an explanation of the information depicted in the plot). The note specifies the number of books that are available in the respective time spans.

To make the inferences less susceptible to random noise, the 1M samples of the time span 1900-1904 were aggregated. The same was done for the samples of the time span 1915-1919. Figure 7 plots the resulting word cloud and six selected time series plots. It becomes immediately obvious that the reason for the observed sudden change is related to orthography: while one-letter words such as the preposition 'a' [*to*] and the conjunctions 'e' [*and* preceding an <i> sound], 'o' [*or*] and 'u' [form of 'o' when preceding an <o> sound] were written with an accent mark until 1910, the accent marks vanished after this time span. Indeed, Marín (1992: 133–134) describes this change of the 'acento ortográfico' conducted by the Spanish Royal Academy, which became effective around 1911. Additionally, the time series plots show that 'i' becomes less frequent while 'y' [*and*] becomes more frequent. To understand this, I used the interactive Google Books search to find phrases containing the one-letter word 'i'.[4] The results indicate that 'i' was used predominantly as a conjunction in Chile (and also in Peru). It was successfully replaced by 'y' around 1910 as part of the otherwise mostly unsuccessful orthography reform of 1844 initiated by Andrés Bello (Collier 2004: 101). The sixth time series plot of Figure 7 shows that 'más' [used as an adverb for *more*] was written without an accent mark up until around 1900, but has been used with an accent mark ever since. These results demonstrate that the method is able to measure linguistic developments that can be linked to historic events.

To demonstrate how the method of measuring corpus similarity might complement existing approaches of natural language processing (NLP), I re-ran the analyses of this section, but also used the available part-of-speech (POS) information. Before explaining the approach, a few notes are in order here: compared to the first release in 2009 (Michel et al. 2010b), the GBC of the second release are not only bigger and have improved OCR and metadata quality, but are also POS tagged (Lin et al. 2012). To train the Conditional Random Field POS tagger, Lin et al. (2012: 171) use 'manually annotated treebanks of modern text (often newswire)'. The language-dependent POS tags are then mapped on the universal POS tagset developed by Petrov et al. (2012) consisting of twelve POS tags (the abbreviations used for the analyses below are given in brackets):

- nouns (N),
- verbs (V),
- adjectives (ADJ),
- adverbs (ADV),
- pronouns (PRO),
- determiners and articles (DET),
- adpositions (ADP)
- numerals (NUM)
- conjunctions (CON)
- particles (PRT)
- punctuation marks (PUNCT) and
- a separate miscellaneous category for other categories such as foreign words or abbreviations (OTH).

Using equation (2), I first extracted the 50 word types that underwent the most pronounced change in frequency. As mentioned above, the POS-tag information was included for this analysis. To detect correlated changes, equation (7) was used to calculate period-to-period changes. Each of those time series for the 50 extracted word types was then correlated with each other using the cosine product as the measure of similarity. The resulting dissimilarity matrix was then used as input for a (weighted average) hierarchical cluster analysis. Figure 8 produces a dendrogram of the cluster analysis. All word types with a plus/minus sign have become more/less frequent in the investigated period of time. Figure 8 shows that the method correctly groups those word types together. The one-letter words that have lost their accent are also categorized as very similar, which effectively means that their period-to-period changes are strongly related.

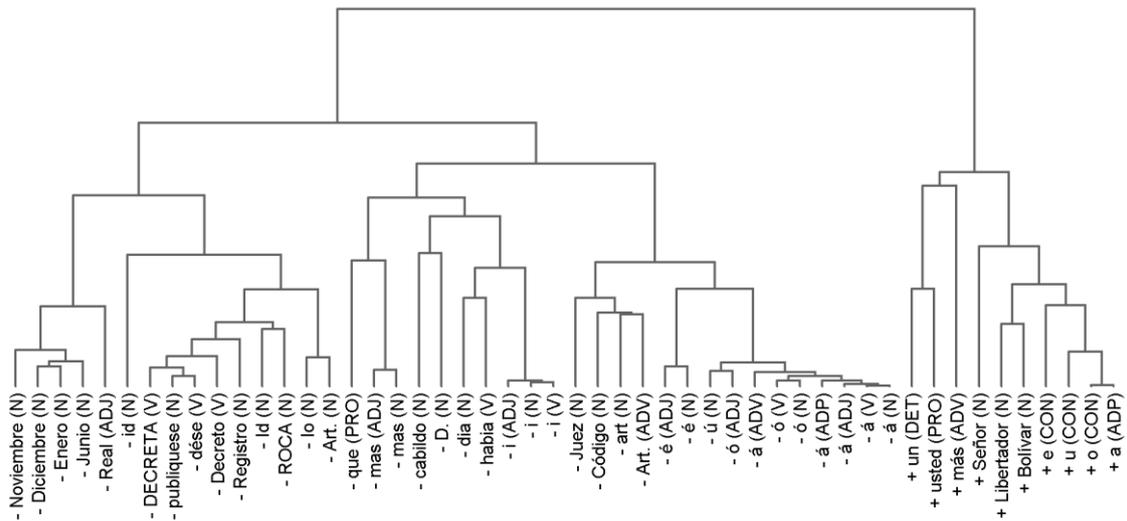

**Figure 8: Dendrogram of the 50 words that underwent the most pronounced change in frequency between the two time spans 1900-1904 and 1915-1920 in the Spanish Google Books Ngram corpus.**

More interestingly, the cluster analysis shows that the word types that have been replaced are incorrectly POS-tagged. Those types are clustered into similar groups. Further analyses (using the same method as in Table 1) indicate that the aggregated frequencies of those time series tend to be strongly negatively correlated with the word types that replaced those word types.

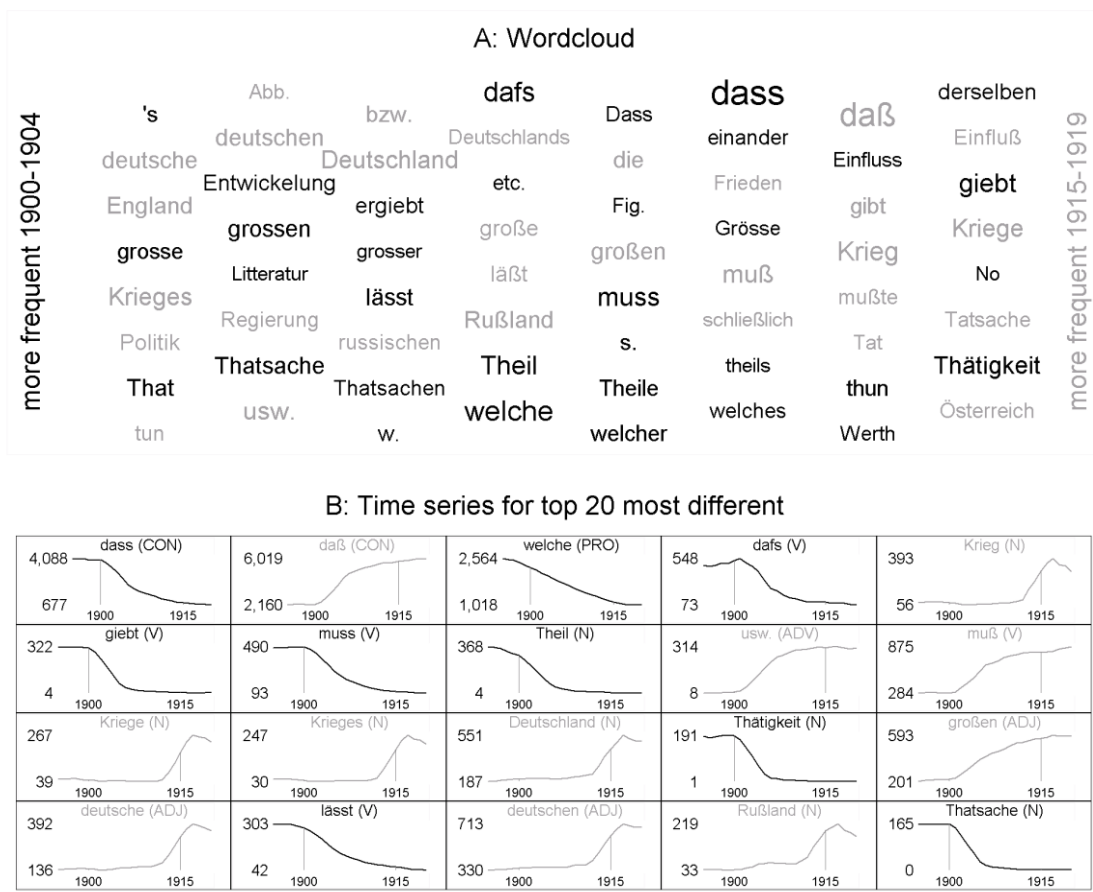

**Figure 9: Visualization of the differences between the time spans 1900-1904 and 1915-1919 in the German Books Ngram data (cf. Figure 5 and Figure 7 for an explanation of the information depicted in the plot).**

Using the bigram GBC and the Google books search indeed shows that, in most cases, the one-letter words which lost their accent mark have been incorrectly classified. In the unigram data 'o' is classified as a conjunction in 100 % of all cases, while 'ó' is only classified as a conjunction in 0.01 % of all instances. Similar ratios can be observed for 'a' which is POS tagged as an adposition in 99.05 % of all cases, while 'á' is only tagged as an adposition in 4.68% of all cases. Finally, 'mas' can mean *farm* in castellan, and would be correctly classified as a noun. However, the bigram data indicates that in most of the cases 'mas' is used as an adverb and is hence incorrectly classified.

This key result is due to the fact that – as mentioned above – the tagger is trained with modern text data. Since especially those one-letter words have a high frequency, 'fine-grained analysis of the evolution of syntax', as Lin et al. (2012: 169) put it, could be seriously biased. Thus, to improve diachronic POS tagging, the method used in this paper might help to select suitable training data for both the POS tagger and parser models (Lin et al. 2012: 171–172). This might be an interesting avenue to investigate in the future.

**German Google Books Ngram lexical changes around 1910**

In a similar vein, the 100 word types (including the POS-tag information) that underwent the most pronounced change in frequency in the German GBC between the time spans 1900-1905 and 1915-1920 were extracted and correlated.

Figure 9 visualizes the change by plotting the 64 word types with the highest partial contribution against the observed difference. Again, small multiples of the 20 most different word types are added. The figure shows that the difference between the two periods of time is the result of two main factors:

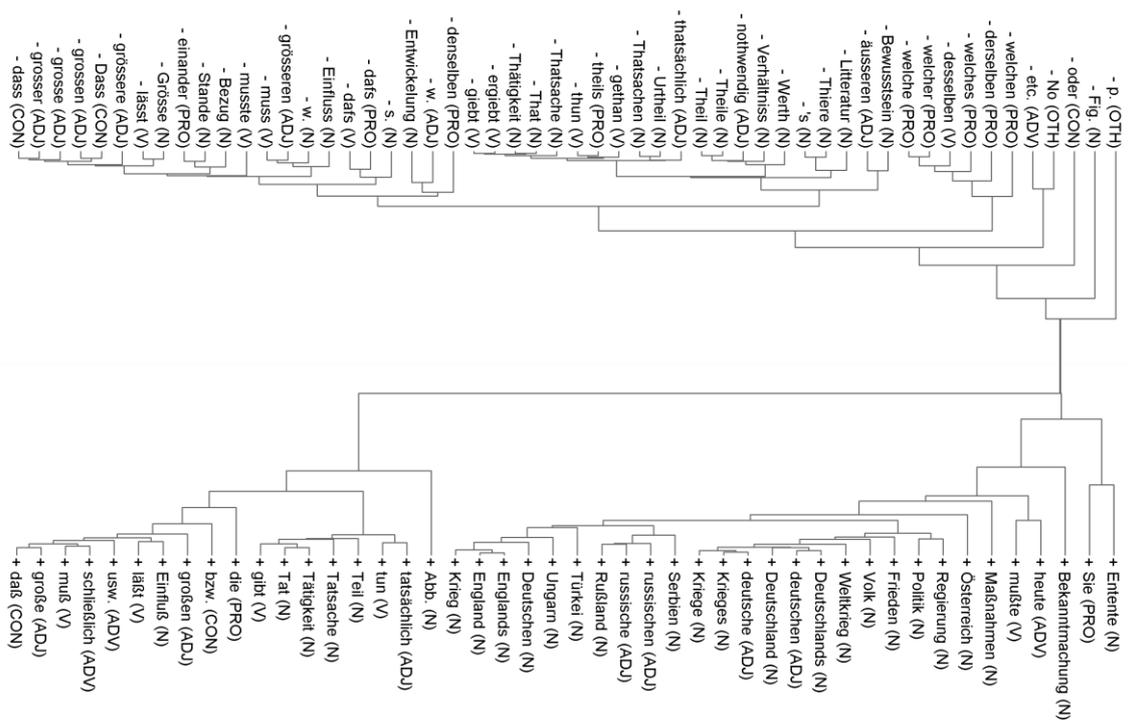

**Figure 10: Dendrogram of the 100 words that underwent the most pronounced change in frequency between the two time spans 1900-1904 and 1915-1920 in the German Google Books Ngram corpus.**

(i) The German orthography reform of 1901, where for example for several words 'th' was replaced by 't' ('Teil' instead of 'Theil' [*part, fraction*]) for several words, 'ie' was replaced by 'i' ('gibt' instead of 'giebt' [form of *to give*]) and it was decided to use 'ß' instead of 'ss' at the end of words or in front of consonants ('muß' instead of 'muss' [form of *to have to*]). Figure 9 perfectly reflects this reform.

(ii) The contingent influences of the First World War. For example the frequency of different forms of the noun 'Krieg' [*war*] rose sharply. Names of participating countries gained in frequency ('Deutschland' [*Germany*],

England, Russland [*Russia*], Österreich [*Austria*]), the same can be seen for the respective adjectives.

Again, the resulting dissimilarity matrix of the 100 word types was used as input for a cluster analysis. Figure 10 shows that the method is able to cluster the two influencing factors mentioned above in a substantial way by generating three groups: on the one hand, we have a cluster (bottom right) that mostly groups together words related to the rise of war. On the other hand, we have two clusters (top, bottom left) that mostly group together words that are affected by the spelling reform.

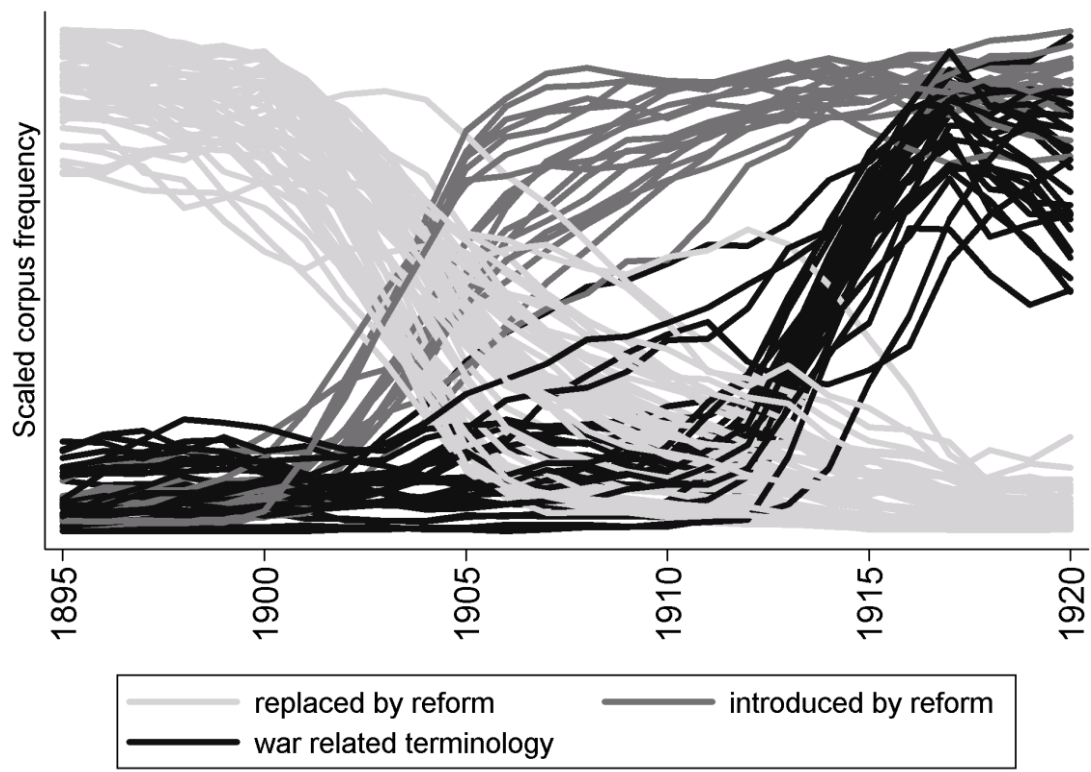

**Figure 11: Scaled corpus frequency for the 100 words that underwent the most pronounced change in frequency between the two time spans 1900-1904 and 1915-1920 in the German Google Books Ngram corpus (scaled to an interval between 0 and 1).**

Additionally, Table 1 lists ten selected pairs of words whose period-to-period correlation are strongly negative and ten selected pairs with a very strong positive correlation. The first three columns of the table demonstrate that period-to-period changes of word types that replaced each other are strongly negatively correlated. The last three columns show word types whose period-to-period changes are strongly positive, including the simultaneous rise of war related terminology and the words affected by the spelling reform that were simultaneously replaced or introduced.

| Strong negative correlation | | | Strong positive correlation | | |
|---|---|---|---|---|---|
| ρ | Word 1 | Word 2 | ρ | Word 2 | Word 2 |
| -0.97 | dass | daß | 0.99 | Thätigkeit | That |
| -0.97 | gibt | giebt | 0.99 | ergiebt | giebt |
| -0.97 | thun | Tätigkeit | 0.99 | Krieges | deutsche |
| -0.96 | That | Tat | 0.98 | Krieges | Krieg |
| -0.96 | gibt | ergiebt | 0.98 | Kriege | deutsche |
| -0.95 | daß | muss | 0.97 | deutsche | deutschen |
| -0.95 | Thätigkeit | Tat | 0.97 | England | Krieg |
| -0.94 | Tat | gethan | 0.96 | Tat | Tätigkeit |
| -0.93 | Thätigkeit | Tätigkeit | 0.95 | Kriege | Weltkrieg |
| -0.91 | Litteratur | Tätigkeit | 0.94 | Rußland | England |

**Table 1: Selected word pairs whose period-to-period is strongly negative (column 1-3) or strongly positive (column 4-6).**

This demonstrates that the approach might also complement novel NLP approaches that try to solve 'typical challenges' of diachronic datasets and thereby are 'enhancing spelling normalization, POS-tagging and subsequent diachronic linguistic analyses' (Amoia and Martínez 2013: 84), by identifying variants of word forms that might have replaced each other, e.g. in order to improve programs developed for normalizing spelling variation in diachronic data (Baron and Rayson 2008).

**Further applications**

To discover 'previously undetected phenomena available for further analysis' (Hilpert and Gries 2009: 398), the reader is invited to visit http://www.owid.de/plus/lc2015/, where it is possible to choose one of seven languages (three varieties of English [American English, British English, English Fiction], French, German, Italian and Spanish), two periods of time from 1800 to 2000 and various part of speech categories. Based on the method presented in this paper, the tool then automatically visualizes the words whose frequency changed most in the GBC for the selected time periods (similar to Figure 5, Figure 7 and Figure 9).[5]

One example that might be interesting for cross cultural and linguistic discourse analyses can be found in the supplemental online material. It demonstrates that the method makes it easy to replicate and extend analyses like the one presented by Pechenick et al. (2015) who argue that the English GBC have seen a rise of scientific and technical literature throughout the second half of the 20th century. Indeed, Figure 12-17 clearly demonstrate that this seems to be not only true for the two English corpora [American and British English] but also for the French, the German, the Italian and the Spanish corpora, as can be seen by an increase of the respective vocabulary (e.

g.'information', 'model', 'data', 'research', 'modèle', 'recherche', 'approche', 'Daten', 'Modell', 'approccio', 'modello', 'ricerca', 'modelo', 'tecnología'). The visualization technique used in this paper makes it also possible to further refine the analysis of Pechenick et al. (2015) who argue that there was a rise of the two *science-related* terms 'percent' and 'Figure'. Figure 18 reveals that 'percent' was written 'per cent' in earlier periods (cf. Koplenig 2015b) and 'Figure' was abbreviated 'Fig.' or 'F.' in the American English GBC.

It is also worth pointing out that in all six GBC corpora, there is a rise of gender related vocabulary ('women', 'gender', 'femmes', 'Frauen', 'donne', 'mujeres', cf. Figure 12-17). This could stimulate further research in historical discourse analysis.

Another example of the use of the method can be found in Koplenig (2015a) where the approach is used to reconstruct the composition of the German corpus in times of World War II (WWII), by demonstrating that the corpus was strongly biased towards volumes published in Switzerland during WWII. This implies that observed changes in this period of time can only be linked directly to World War II to a certain extent and therefore questions the idea of directly measuring censorship in Nazi Germany based on the GBC data (Michel et al. 2010a).

## 8. Concluding remarks

In this paper, a method to measure synchronic corpus similarity put forward by Kilgarriff (2001) was adapted and extended to identify trends and correlated changes in diachronic corpus data. It was shown that the method is both intuitively plausible and motivated in terms of a concept from information theory. It even has a natural interpretation for diachronic data, as it extracts time series that underwent the most pronounced trend in a given time span. This makes it possible to use the method in the context of other temporally ordered quantities. The main advantages of this exploratory approach are that (i) it is computationally very cheap, (ii) it works very fast, (iii) it is fully data-driven and (iv) it only needs type and token frequency lists as input. For the synchronic case study, it was shown that the method can automatically identify unigrams with the largest individual contribution to the observed (dis-)similarity that are typical for that which the different corpora are designed to represent, e.g. interactive or involved text production.

For the diachronic case study, those lists were used to extract candidate words that changed most during a given period of time. These word types were used to detect correlated changes. It was demonstrated that the method could help to improve diachronic POS tagging and to understand lexical developments that can be linked to historical events.

This does not imply that the method is restricted to unigram frequencies. It could easily be applied to n-grams of any order or other quantifiable linguistic structures, such as changing verb productivity patterns. As Hilpert points out, 'many theoretical questions do no straightforwardly relate to frequency' (2011: 457), and equally 'important measures concern changes in the interdependencies of one linguistic unit with other units in its environment' (2011: 442). It may even be possible to extend this classical view of collocation, as the special properties of time series make it possible to analyze evolving relationships without being restricted to adjacent sequences of words or terms, so that complex co-occurrence patterns across different texts could be analyzed (cf. Koplenig 2015b for a first step in this direction). This relationship does not even have to evolve in equal periods but can be shifted in time and then measured by the cross-correlation of the involved time series (Becketti 2013: 330–335). Furthermore, correlated changes across different languages could potentially be measured.

On the downside, the sheer amount of available data also imposes some methodological restrictions on the approach. Some period-to-period changes of paired time series will sometimes look highly correlated just as the result of chance. Developing a method that is not hampered by such spurious correlations is an important avenue for future research.

To sum up, I hope that I have demonstrated that the method presented in this paper is able to incorporate a suggestion made by Hilpert (2011: 438), as it outlines a technique that is 'both adapted towards the special characteristics of diachronic corpus data [and is] designed to address [some] types of theoretical questions that matter to linguists interested in diachronic change for a first step in this direction. While always keeping in mind that historical linguistics can 'be thought of as the art of making the best use of bad data. for a first step in this direction (Labov 1994: 11), the method may complement other existing data driven approaches such as the one by Gries and Hilpert (2008), or it may help to extract interesting patterns that can then be used as input for dynamical visualizations of linguistic changes, e.g. building on motion charts, as presented by Hilpert (2011).

**Acknowledgments**

I would like to thank Stefan Engelberg, Martin Hilpert, Alexander Hinneburg, Marc Kupietz, Peter Meyer, Frank Michaelis, Sarah Signer, Carolin Müller-Spitzer, Antje Töpel and Sascha Wolfer for many helpful discussions and for (proof-)reading a draft version of this paper. All remaining errors are mine.

**Notes:**

[1] Since this process is random *per definition*, instead of using 1,000,000 as the nominator for $N_c$, 1,010,000 million was used to obtain a sample which is slightly bigger than 1,000,000 tokens. To generate a sample of exactly 1,000,000 tokens, all drawn tokens were thrown in an urn from which 1,000,000 million tokens were then drawn randomly.

[2] For the sampling procedure, tokens tagged as numerals and punctuation were excluded to account for some obvious errors in the tokenization and tagging of POS (Michel et al. 2010b). Words that were longer than five characters and did not contain at least one alphanumeric character (regular expression: [A-Za-z0-9]) were excluded (e.g. ****** , ...... , ------, ______). Strings consisting solely of the following characters were removed, too: « » . ' * § • .. ° # $.+^* ( ) [ ] { } - = | \ : ; < , > ? / ~ ` for a first step in this direction. Finally, words consisting of only numeric characters were excluded.

[3] Content words were defined as belonging to one of the following CLAWS POS category (cf. http://www.kilgarriff.co.uk/BNClists/poscodes.html; last accessed 21.Oktober 2014): aj0, aj0-av0, aj0-nn1, aj0-vvd, aj0-vvg, ajc, ajs, av0, nn0, nn1, nn1-np0, nn1-vvb, nn1-vvg, nn2, nn2-vvz, np0, vvb, vvd, vvd-vvn, vvg, vvi, vvn, vvz. The rest of the word types were flagged as function words. The analysis was conducted separately for each category.

[4] Last accessed on 31 October 2014. It is important to keep in mind, however, that not all books that are available via the Google Books search are also included in the GBC. Only books whose OCR quality reaches a certain threshold are included (Michel et al. 2010b).

[5] I would like to thank Frank Michaelis for programming the graphical user interface.

# References

Amoia, Marilisa & José Manuel Martínez. 2013. Using Comparable Collections of Historical Texts for Building a Diachronic Dictionary for Spelling Normalization. *Proceedings of the 7th Workshop on Language Technology for Cultural Heritage, Social Sciences, and Humanities*, 84–89. Sofia, Bulgaria: Association for Computational Linguistics. http://www.aclweb.org/anthology/W13-2711.

Baron, Alistair & Paul Rayson. 2008. VARD 2: A tool for dealing with spelling variation in historical corpora. *Proceedings of the Postgraduate Conference in Corpus Linguistics*. Aston University, Birmingham, UK. http://acorn.aston.ac.uk/conf_proceedings.html (25 August, 2015).

Becketti, Sean. 2013. *Introduction to time series using Stata*. 1st ed. College Station, Tex: Stata Press.

Bentz, Christian, Douwe Kiela, Felix Hill & Paula Buttery. 2014. Zipf's law and the grammar of languages: A quantitative study of Old and Modern English parallel texts. *Corpus Linguistics and Linguistic Theory* 10(2). doi:10.1515/cllt-2014-0009.

Biber, Douglas & Edward Finegan. 1989. Drift and the Evolution of English Style: A History of Three Genres. *Language* 65(3). 487. doi:10.2307/415220 (1 July, 2014).

Bochkarev, Vladimir, Valery Solovyev & Sören Wichmann. 2014. Universals versus historical contingencies in lexical evolution. http://wwwstaff.eva.mpg.de/%7Ewichmann/LexEvolUploaded.pdf (12 June, 2014).

Burnard, Lou (ed.). 2007. [bnc] British National Corpus. http://www.natcorp.ox.ac.uk/docs/URG/ (21 October, 2014).

Collier, Simon. 2004. *A history of Chile, 1808-2002*. 2nd. ed. (Cambridge Latin American Studies 82). Cambridge [England] ; New York, N.Y: Cambridge University Press.

Cox, N. J. 2012. Speaking Stata: Transforming the time axis. *Stata Journal* 12(2). 332–341.

Cramér, Harald. 1946. *Mathematical methods of statistics*. Princeton: Princeton University Press.

Culturomics. 2014. www.culturomics.org. *www.culturomics.org*. http://www.culturomics.org/ (8 September, 2014).

Davies, Mark. 2007. TIME Magazine corpus: 100 million words, 1820s-2000s. http://corpus.byu.edu/time (16 October, 2014).

Davies, Mark. 2010a. The Corpus of Historical American English: 400 million words, 1810-2009. http://corpus.byu.edu/coha/ (16 October, 2014).

Davies, Mark. 2010b. The Corpus of Historical American English: COMPOSITION OF THE CORPUS. http://corpus.byu.edu/coha/files/cohaTexts.xls (24 October, 2014).

Davies, Mark. 2012. Expanding horizons in historical linguistics with the 400-million word Corpus of Historical American English. *Corpora* 7(2). 121–157. doi:10.3366/cor.2012.0024 (14 January, 2015).

Davies, Mark. 2014. Making Google Books n-grams useful for a wide range of research on language change. *International Journal of Corpus Linguistics* 19(3). 401–416.

Endres, D.M. & J.E. Schindelin. 2003. A new metric for probability distributions. *IEEE Transactions on Information Theory* 49(7). 1858–1860. doi:10.1109/TIT.2003.813506 (17 October, 2014).

Granger, C.W.J. & P. Newbold. 1974. Spurious regressions in econometrics. *Journal of Econometrics* 2(2). 111–120. doi:10.1016/0304-4076(74)90034-7 (23 June, 2014).

Gries, Stefan Th. & Martin Hilpert. 2008. The identification of stages in diachronic data: variability-based neighbor clustering. *Corpora* 3(1). 59–81.

Hilpert, Martin. 2011. Dynamic visualizations of language change: Motion charts on the basis of bivariate and multivariate data from diachronic corpora. *International Journal of Corpus Linguistics* 16(4). 435–461. doi:10.1075/ijcl.16.4.01hil (17 June, 2014).

Hilpert, Martin & Stefan Th. Gries. 2009. Assessing frequency changes in multistage diachronic corpora: Applications for historical corpus linguistics and the study of language acquisition. *Literary and Linguistic Computing* 24(4). 385–401. doi:10.1093/llc/fqn012.

Huber, Magnus, Magnus Nissel, Patrick Maiwald & Bianca Widlitzki. 2012. The Old Bailey Corpus. Spoken English in the 18th and 19th centuries. www.uni-giessen.de/oldbaileycorpus (15 January, 2015).

Hughes, James M., Nicholas J. Foti, David C. Krakauer & Daniel N. Rockmore. 2012. Quantitative patterns of stylistic influence in the evolution of literature. *Proceedings of the National Academy of Sciences* 109(20). 7682–7686. doi:10.1073/pnas.1115407109 (10 March, 2014).

Jurafsky, Daniel & James H Martin. 2009. *Speech and Language processing: an introduction to natural language processing, computational Linguistics, and speech recognition*. Upper Saddle River: Pearson Education (US).

Kilgarriff, Adam. 1997. Putting frequencies in the dictionary. *International Journal of Lexicography* 10(2). 135–155.

Kilgarriff, Adam. 2001. Comparing Corpora. *International Journal of Corpus Linguistics* 6(1). 97–133. doi:10.1075/ijcl.6.1.05kil (19 May, 2014).

Kilgarriff, Adam. 2012. Getting to Know Your Corpus. In Petr Sojka, Aleš Horák, Ivan Kopeček & Karel Pala (eds.), *Text, Speech and Dialogue*, vol. 7499, 3–15. Berlin, Heidelberg: Springer Berlin Heidelberg. http://link.springer.com/10.1007/978-3-642-32790-2_1 (2 February, 2015).

Kilgarriff, Adam. 2014. Read-me for Kilgarriff's BNC word frequency lists. http://www.kilgarriff.co.uk/bnc-readme.html (20 October, 2014).

Kilgarriff, Adam & Raphael Salkie. 1996. Corpus Similarity and Homogeneity via Word Frequency. *Euralex '96 proceedings: papers submitted to the Seventh EURALEX International Congress on Lexicography in Göteborg, Sweden. 1. 1.*, 121–130. Göteborg: Univ., Dep. of Swedish.

Klingenstein, S., T. Hitchcock & S. DeDeo. 2014. The civilizing process in London's Old Bailey. *Proceedings of the National Academy of Sciences* 111(26). 9419–9424. doi:10.1073/pnas.1405984111 (17 October, 2014).

Köhler, Reinhard. 2012. *Quantitative syntax analysis*. (Quantitative Linguistics 65). Berlin: De Gruyter Mouton.

Kohnen, Thomas. 2007. From Helsinki through the centuries: The design and development of English diachronic corpora." In: Towards Multimedia in Corpus Studies. In Päivi Pahta, Irma Taavitsainen, Terttu Nevalainen & Jukka Tyrkkö (eds.), *Helsinki: Research Unit for Variation, Contacts and Change in English.* (Studies in Language Variation, Contacts and Change in English 2). http://www.helsinki.fi/varieng/journal/volumes/02/kohnen (5 October, 2014).

Koplenig, Alexander. 2015a. The impact of lacking metadata for the measurement of cultural and linguistic change using the Google Ngram datasets – reconstructing the composition of the German corpus in times of WWII. *Digital Scholarship in the Humanities*.

Koplenig, Alexander. 2015b. Using the parameters of the Zipf–Mandelbrot law to measure diachronic lexical, syntactical and stylistic changes – a large-scale corpus analysis. *Corpus Linguistics and Linguistic Theory* 0(0). doi:10.1515/cllt-2014-0049. http://www.degruyter.com/view/j/cllt.ahead-of-print/cllt-2014-0049/cllt-2014-0049.xml (19 April, 2015).

Koplenig, Alexander. 2015c. Why the quantitative analysis of diachronic corpora that does not consider the temporal aspect of time-series can lead to wrong conclusions. *Digital Scholarship in the Humanities*. fqv030. doi:10.1093/llc/fqv030.

Labov, William. 1994. *Principles of linguistic change*. (Language in Society 20). Oxford, UK ; Cambridge [Mass.]: Blackwell.

Lin, Yuri, Jean-Baptiste Michel, Lieberman Erez Aiden, Jon Orwant, Will Brockmann & Slav Petrov. 2012. Syntactic Annotations for the Google Books Ngram Corpus. *Proceedings of the 50th Annual Meeting of the Association for Computational Linguistics*, 169–174. Jeju, Republic of Korea.

Marín, Juan Martínez. 1992. La ortografía española: perspectiva historiogáfica. *Cauce* 14-15. 125–134.

Michel, Jean-Baptiste, Yuan Kui Shen, Aviva Presser Aiden, Adrian Verses, Matthew K. Gray, The Google Books Team, Joseph P. Pickett, et al. 2010a. Quantitative Analysis of Culture Using Millions of Digitized Books. *Science* 331(14). 176–182 [online pre–print: 1–12]. doi:10.1126/science.1199644.

Michel, Jean-Baptiste, Yuan Kui Shen, Aviva Presser Aiden, Adrian Verses, Matthew K. Gray, The Google Books Team, Joseph P. Pickett, et al. 2010b. Quantitative Analysis of Culture Using Millions of Digitized Books (Supporting Online Material). *Science* 331(14). doi:10.1126/science.1199644. http://www.sciencemag.org/content/early/2010/12/15/science.1199644/suppl/DC1 (5 March, 2014).

Mota, Cristina. 2010. Journalistic Corpus Similarity over Time. In Stefan Thomas Gries, Stefanie Wulff & Mark Davies (eds.), *Corpus-linguistic applications: current studies, new directions*, 67–84. (Language and Computers no. 71). Amsterdam ; New York: Rodopi.

Pechenick, Eitan Adam, Christopher M. Danforth & Peter Sheridan Dodds. 2015. Characterizing the Google Books corpus: Strong limits to inferences of socio-cultural and linguistic evolution. http://arxiv.org/abs/1501.00960.

Petrov, Slav, Dipanjan Das & Ryan McDonald. 2012. A Universal Part-of-Speech Tagset. In Nicoletta Calzolari (Conference Chair), Khalid Choukri, Thierry Declerck, Mehmet Uğur Doğan, Bente Maegaard, Joseph Mariani, Asuncion Moreno, Jan Odijk & Stelios Piperidis (eds.), *Proceedings of the Eighth International Conference on Language Resources and Evaluation (LREC'12)*. Istanbul, Turkey: European Language Resources Association (ELRA).

Piantadosi, Steven T. 2014. Zipf's word frequency law in natural language: A critical review and future directions. *Psychonomic Bulletin & Review*. doi:10.3758/s13423-014-0585-6. http://link.springer.com/10.3758/s13423-014-0585-6 (2 May, 2014).

Rissanen, Matti, Merja Kytö, Leena Kahlas-Tarkka, Matti Kilpiö, Saara Nevanlinna, Irma Taavitsainen, Terttu Nevalainen & Helena Raumolin-Brunberg. 1991. The Helsinki Corpus of English Texts. Department of Modern Languages , University of Helsinki. Compiled by Matti Rissanen (Project leader), Merja Kytö (Project secretary); Leena Kahlas-Tarkka, Matti Kilpiö (Old English); Saara Nevanlinna, Irma Taavitsainen (Middle English); Terttu Nevalainen, Helena Raumolin-Brunberg (Early Modern English). http://www.helsinki.fi/varieng/CoRD/corpora/HelsinkiCorpus/ (25 August, 2015).

Tufte, Edward R. 2001. *The visual display of quantitative information*. 2nd ed. Cheshire, Conn: Graphics Press.

Tweedie, Fiona J. & R. Harald Baayen. 1998. How Variable May a Constant be? Measures of Lexical Richness in Perspective. *Computers and the Humanities* 32(5). 323–352.

**Supplemental online material**

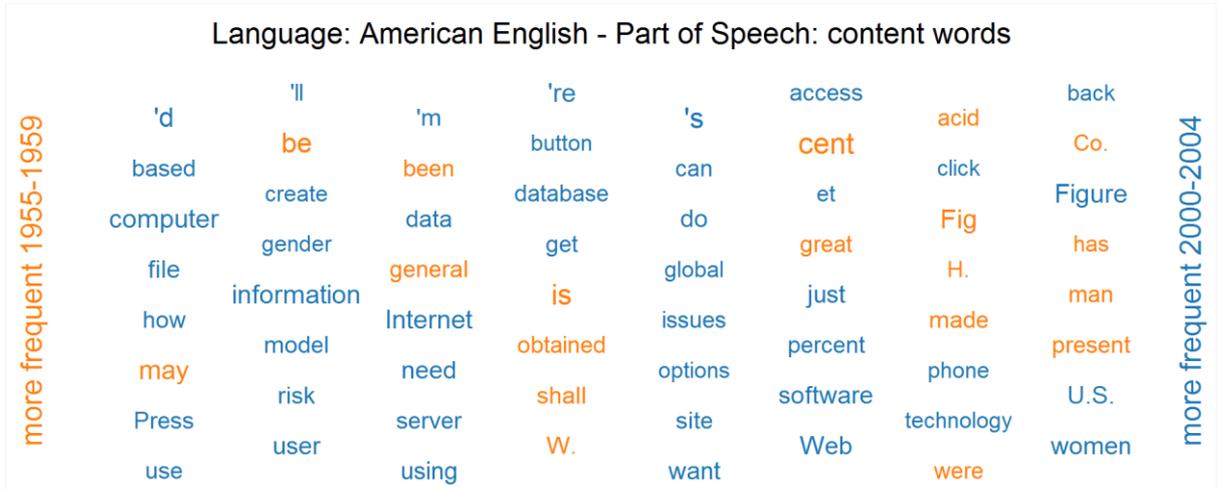

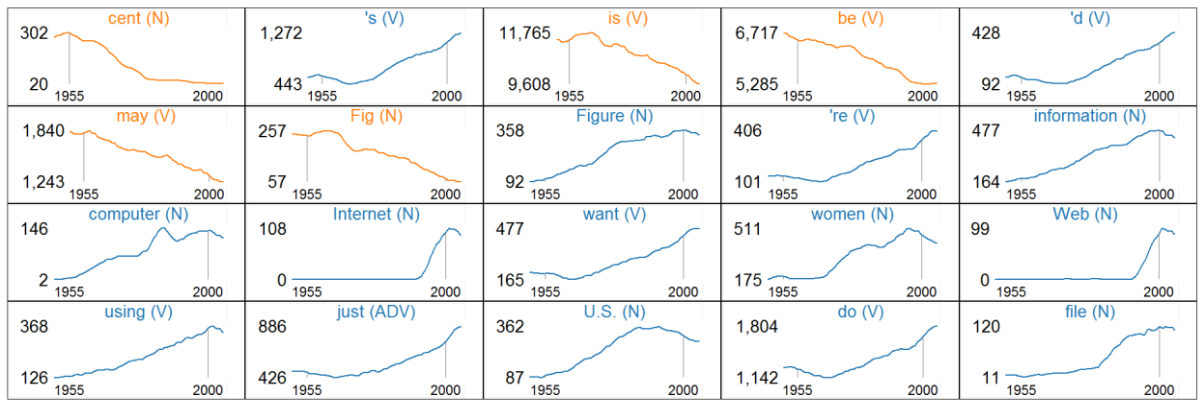

**Figure 12: Visualization of the differences between the time spans 1955-1959 and 2000-2004 in the American English GBC; source: http://www.owid.de/plus/lc2015/**

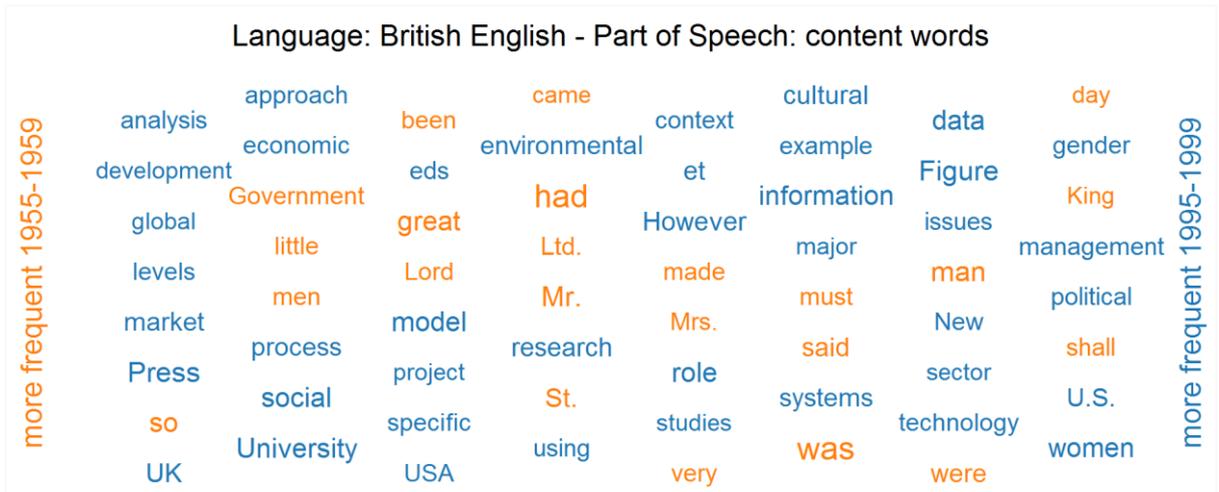

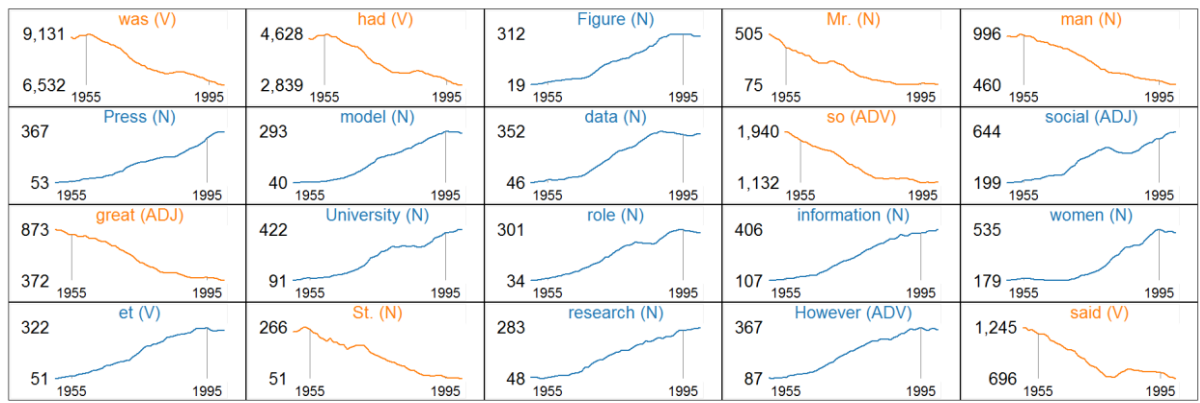

**Figure 13: Visualization of the differences between the time spans 1955-1959 and 1995-1999 in the British English GBC; source: http://www.owid.de/plus/lc2015/**

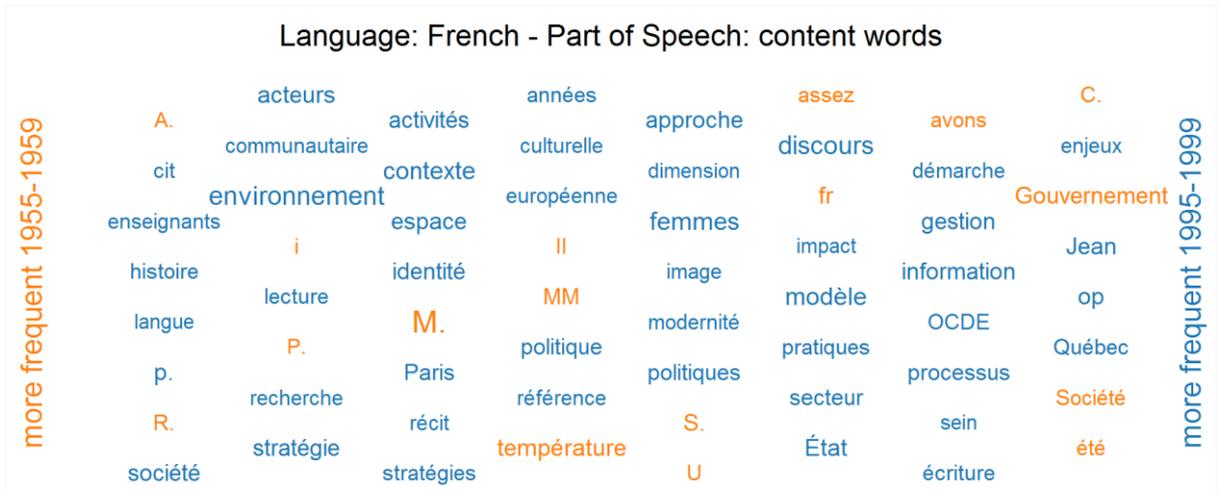

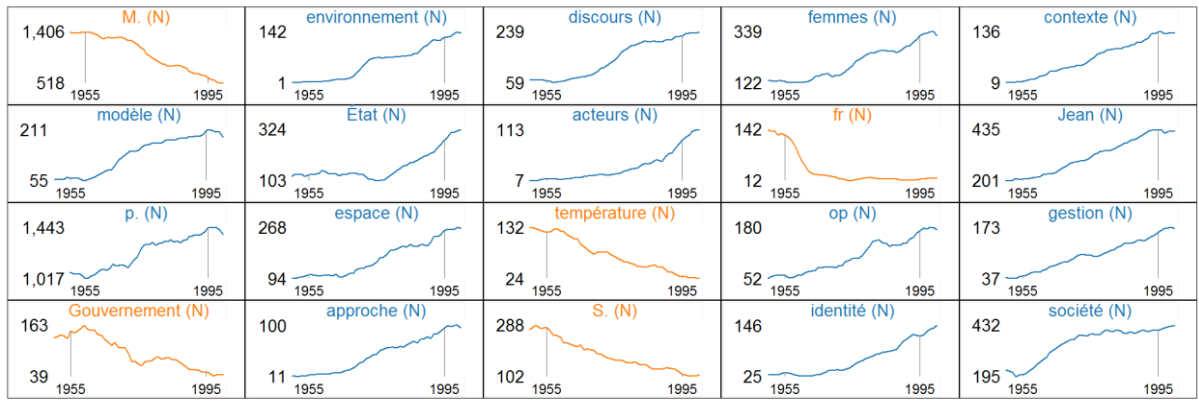

**Figure 14: Visualization of the differences between the time spans 1955-1959 and 1995-1999 in the French GBC; source: http://www.owid.de/plus/lc2015/**

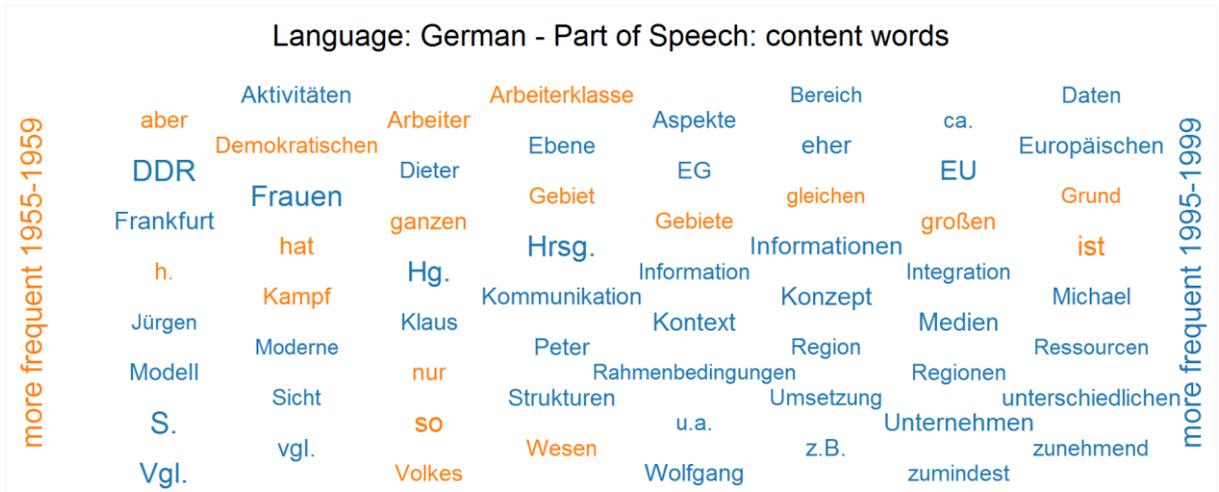

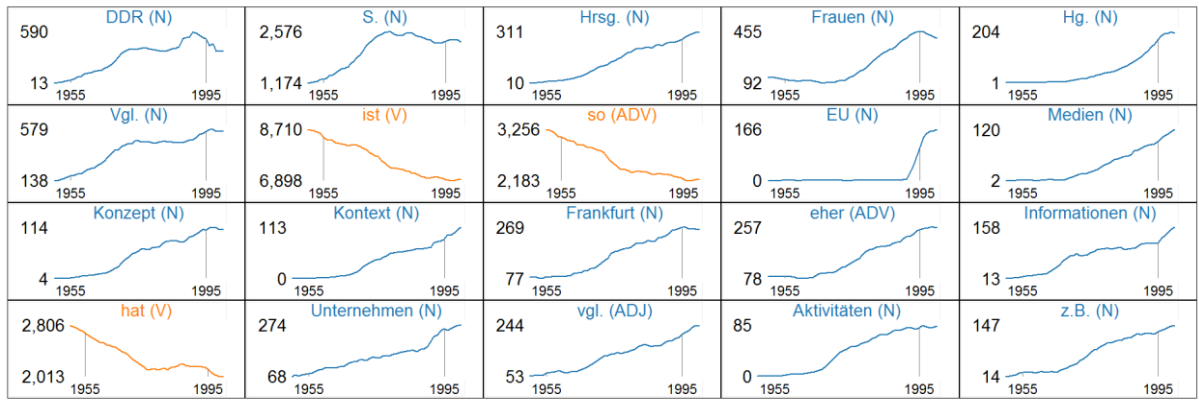

**Figure 15: Visualization of the differences between the time spans 1955-1959 and 1995-1999 in the German GBC; source: http://www.owid.de/plus/lc2015/**

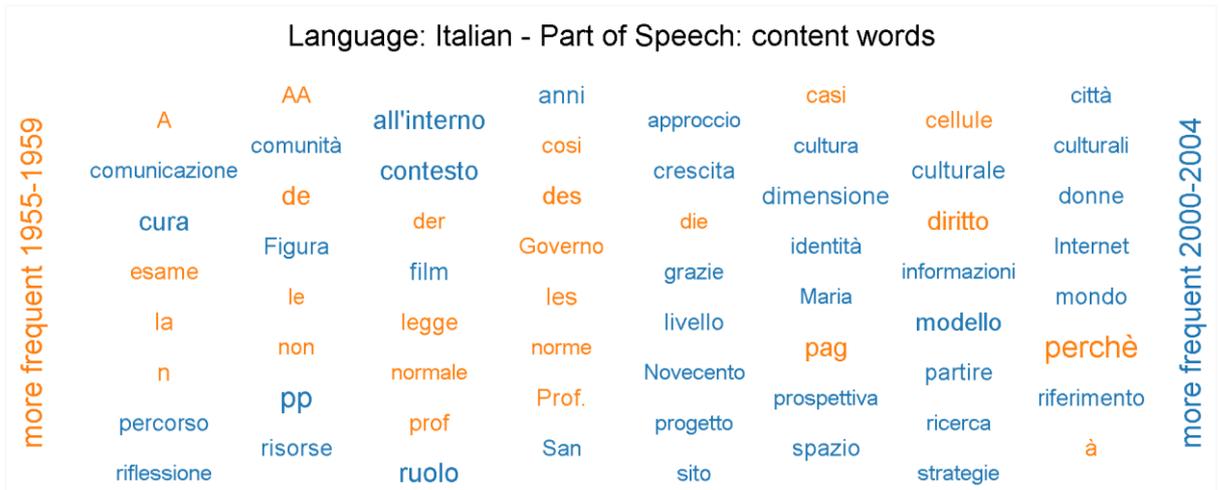

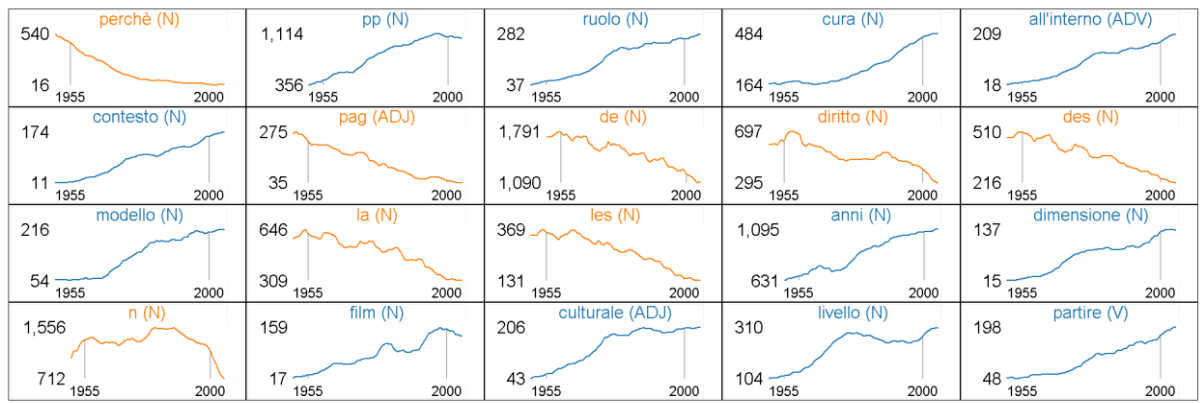

**Figure 16: Visualization of the differences between the time spans 1955-1959 and 2000-2004 in the Italian GBC; source: http://www.owid.de/plus/lc2015/**

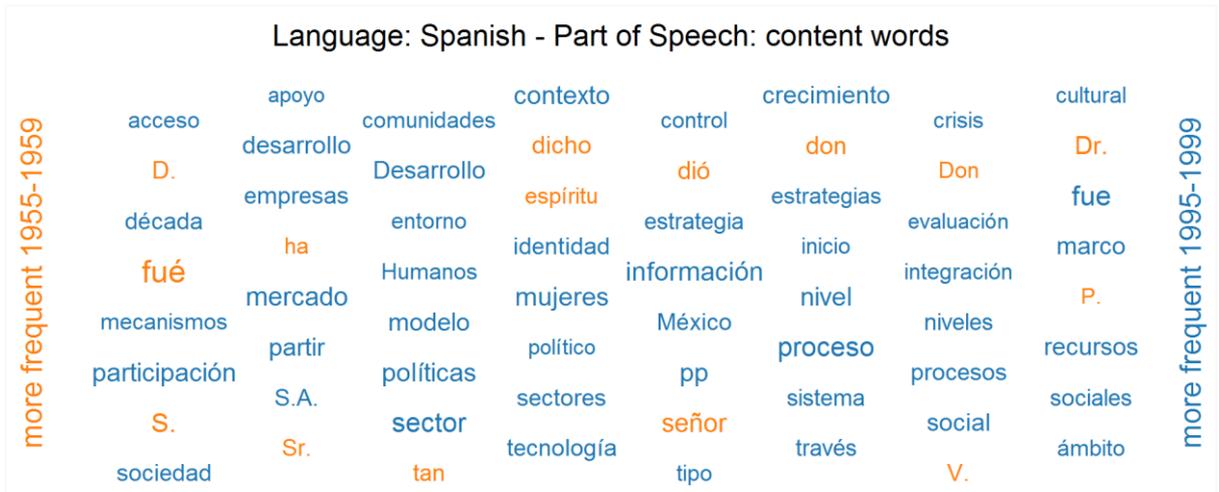

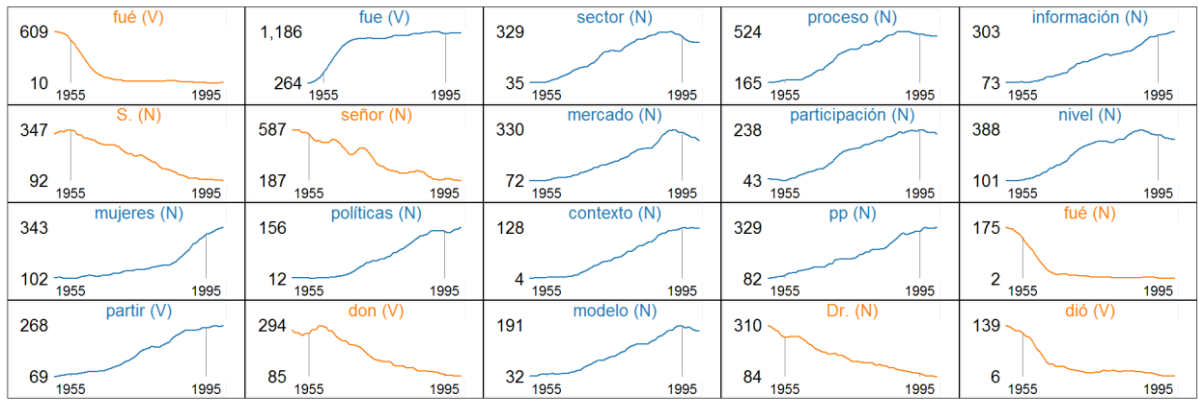

**Figure 17: Visualization of the differences between the time spans 1955-1959 and 1995-1999 in the Spanish GBC; source: http://www.owid.de/plus/lc2015/**

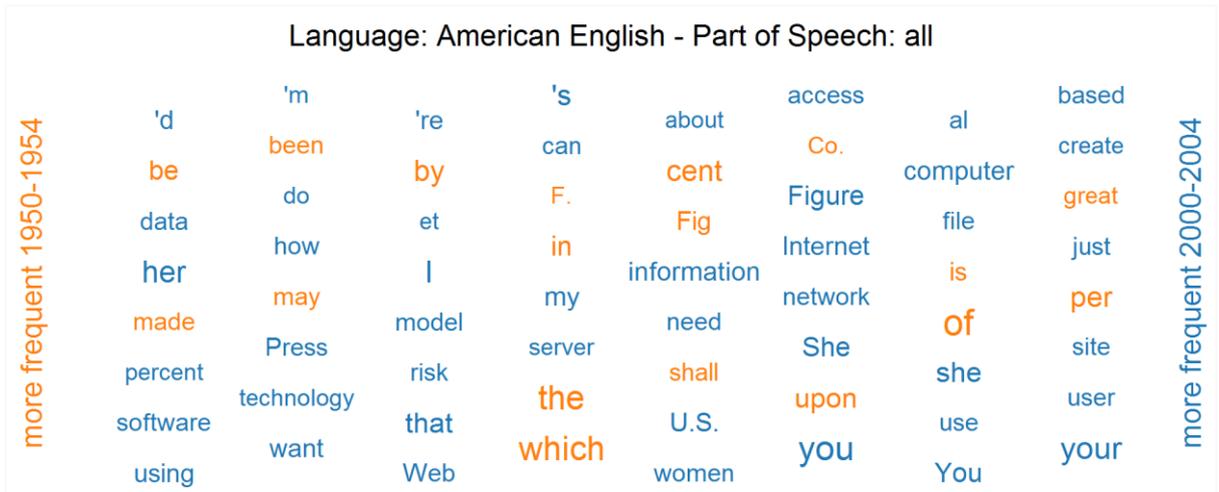

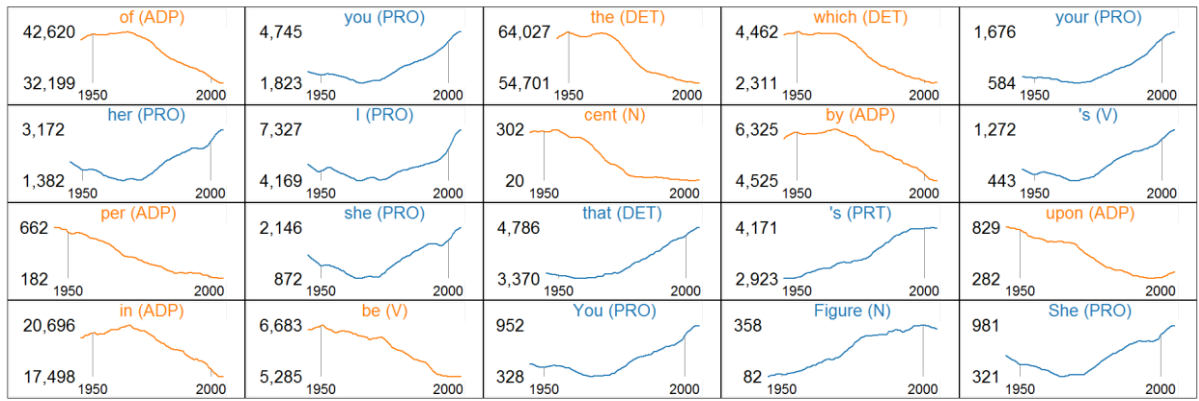

**Figure 18: Visualization of the differences between the time spans 1950-1954 and 2000-2004 in the British English GBCN; source: http://www.owid.de/plus/lc2015/**